\documentclass[conference]{IEEEtran}
\IEEEoverridecommandlockouts
\usepackage{cite}
\usepackage{amsmath,amssymb,amsfonts}
\usepackage{multicol}
\usepackage{algorithmic}
\usepackage{multirow}
\usepackage{enumerate}
\usepackage{etoolbox}
\usepackage{amsmath,graphicx}
\usepackage{lipsum}
\usepackage{bbm}
\usepackage{booktabs}
\usepackage{dsfont}
\usepackage{yfonts}
\usepackage{amssymb}
\usepackage{graphicx}
\usepackage{mathrsfs}
\usepackage{textcomp}
\usepackage{cleveref}
\usepackage{xcolor}
\usepackage{makecell}
\usepackage{float}
\usepackage{subfig}
\def\BibTeX{{\rm B\kern-.05em{\sc i\kern-.025em b}\kern-.08em
    T\kern-.1667em\lower.7ex\hbox{E}\kern-.125emX}}
\begin{document}

\title{Robust Watermarking for Video Forgery Detection with Improved Imperceptibility and Robustness\\
\thanks{This work is supported by National Natural Science Foundation of China under Grant U20B2051, U1936214. $^{\star}$ Corresponding author: Zhenxing Qian (zxqian@fudan.edu.cn).}
}

\author{
\IEEEauthorblockN{Yangming Zhou}
\IEEEauthorblockA{\textit{School of Computer Science} \\
\textit{Fudan University, Shanghai, China}\\
ymzhou21@m.fudan.edu.cn}
\and
\IEEEauthorblockN{Qichao Ying}
\IEEEauthorblockA{\textit{School of Computer Science} \\
\textit{Fudan University, Shanghai, China}\\
shinydotcom@163.com}
\and
\IEEEauthorblockN{Xiangyu Zhang}
\IEEEauthorblockA{\textit{School of Computer Science} \\
\textit{Fudan University, Shanghai, China}\\
xyzhang20@fudan.edu.cn}
\and
\IEEEauthorblockN{Zhenxing Qian$^{\star}$}
\IEEEauthorblockA{\textit{School of Computer Science} \\
\textit{Fudan University, Shanghai, China} \\
zxqian@fudan.edu.cn}
\and
\IEEEauthorblockN{Sheng Li}
\IEEEauthorblockA{\textit{School of Computer Science} \\
\textit{Fudan University, Shanghai, China} \\
lisheng@fudan.edu.cn}
\and
\IEEEauthorblockN{Xinpeng Zhang}
\IEEEauthorblockA{\textit{School of Computer Science} \\
\textit{Fudan University, Shanghai, China} \\
zhangxinpeng@fudan.edu.cn}
}

\maketitle
\definecolor{green}{rgb}{0, 0.5, 0}
\definecolor{orange}{rgb}{0.8, 0.6, 0.2}
\definecolor{orange2}{rgb}{1.0, 0.6, 0.2}
\definecolor{red}{rgb}{1.0, 0.0, 0.0}
\definecolor{blue}{rgb}{0.0, 0.0, 1.0}
\definecolor{teal}{rgb}{0.0, 0.4, 0.4}
\definecolor{purple}{rgb}{0.65,0,0.65}
\definecolor{saffron}{rgb}{0.95,0.75,0.2}
\definecolor{turquoise}{rgb}{0.0,0.5,0.5}
\definecolor{black}{rgb}{0.0, 0.0, 0.0}
\definecolor{gray}{rgb}{0.5, 0.5, 0.5}

\newcommand{\bluemarker}[1]{{\color{blue}#1}}
\newcommand{\redmarker}[1]{{\color{red}#1}}
\newcommand{\qcying}[1]{{\color{turquoise}#1}}
\newcommand{\cg}[1]{{\color{black}#1}}

\begin{abstract}
Videos are prone to tampering attacks that alter the meaning and deceive the audience. Previous video forgery detection schemes find tiny clues to locate the tampered areas. 
However, attackers can successfully evade supervision by destroying such clues using video compression or blurring. 
This paper proposes a video watermarking network for tampering localization. 
We jointly train a 3D-UNet-based watermark embedding network and a decoder that predicts the tampering mask. The perturbation made by watermark embedding is close to imperceptible.
Considering that there is no off-the-shelf differentiable video codec simulator, we propose to mimic video compression by ensembling simulation results of other typical attacks, e.g., JPEG compression and blurring, as an approximation. Experimental results demonstrate that our method generates watermarked videos with good imperceptibility and robustly and accurately locates tampered areas within the attacked version.
\end{abstract}


\section{Introduction}
With the maturity of various video processing and compression technologies, the Online Social Networks (OSNs) are crowded with daily-shared videos for entertainment and reporting. 
However, the popularization of video technology also breeds malicious even illegal activities such as the generation of fake news caused by video clipping or tampering.
Manual video inspection and anomaly detection are time-consuming, labor-intensive and usually with high latency.
Therefore, algorithm-based automatic video tampering detection has gained extensive research interest.

Traditional video forgery detection methods~\cite{shelke2021comprehensive,subramanyam2012video,aloraini2020sequential} include stitching detection, copy-paste detection, image restoration detection, etc. 
For example, Subramanyam et al.~\cite{subramanyam2012video} finds that HoG features are robust against various signal processing manipulations. 
Aloraini~\cite{aloraini2020sequential} performs sequential analysis by modeling video sequences as stochastic processes. Changes in the parameters of these processes indicate a video forgery.
In recent years, most of the video forgery detection work focuses on the detection of forged faces~\cite{li2018exposing,li2018learning,zhang2019detecting,zhang2022patch}. 
Li et al.~\cite{li2018learning} considers both temporal and spatial information, and uses 3DCNN to discriminate forged videos. 
Zhang et al.~\cite{zhang2019detecting} mined some traces in the frequency spectrum and detected the images generated by GAN in the frequency domain.
There are also several image manipulation detection schemes. 
Wu et al.~\cite{wu2019mantra} proposed an end-to-end deep neural network structure, ManTraNet, which first learns image manipulation traces through self-supervision, then extracts local anomalies through Z-score, and detects and locates multiple tampering by judging anomalies. 
Dong et al.~\cite{dong2021mvss} proposed MVSS-Net to augment the differences between the tampered and untampered regions at the boundary, and noise inconsistency and edge supervision are monitored to unveil image manipulation. 
However, universal video tampering detection is still a hard issue. One reason is that the above methods either focus on a typical distribution of videos, such as facial clips, or cannot generalize well on compressed videos. Another reason is that video post-processing attacks represented by MPEG compression are complicated, and the ways of video tampering and post-processing are indefinite. Thus, it is extremely difficult to find a universal clue for all kinds of tampered videos.

\begin{figure*}[t]
\centering
\includegraphics[width=1.0\linewidth]{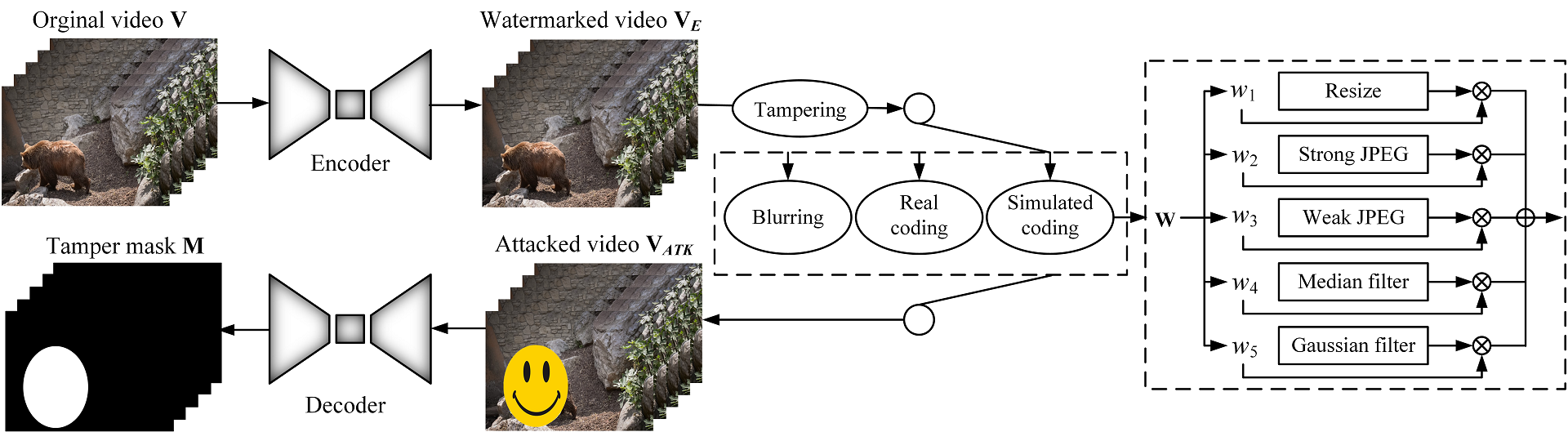}
\caption{\textbf{Sketch of the pipeline of RWVMD.} A 3D-UNet based encoder is used to embed the watermark information into the video. Video processing attacks in social media are simulated by the attack layer. Finally, the decoder predicts the tampering mask from the watermarked video.}
\label{Fig1}
\end{figure*}
Active forensics based on watermarking is an important alternative for manipulation detection. The goal is to hide a tailored clue into the targeted videos for protection, and once the embedded signal is destroyed by tampering, the recipient can identify the modified areas. Meanwhile, the signal must survive video post-processing attacks to ensure robustness. 
In the image domain, Imuge~\cite{ying2021image} presents a robust watermarking scheme that protects images from being tampered with. After the recipient gets the tampered protected image, he can conduct accurate tamper localization and image recovery. 
Khachaturov~\cite{khachaturov2021markpainting} proposes a watermarking-like adversarial method, called Markpainting that prevents images from being inpainted.
Besides, many video watermarking methods~\cite{ying2019robust,wang2022dtcwt,asikuzzaman2015blind} have been proposed for covert data transmission.
But they only focus on hiding as much information as possible into a targeted video, and in comparison, there is little work that focuses on video watermarking against tampering attacks. 

We propose an end-to-end Robust Watermarking network for Video Forgery Detection (RWVFD). We jointly train a 3D-UNet-based encoder for imperceptible watermark embedding and a  decoder for tampering localization.  We design an attack simulation module that combines simulated video encoding, real video encoding, and other obfuscation attacks to improve the robustness of watermarking. Considering that there is no off-the-shelf differentiable video codec simulator, we propose to mimic video compression by ensembling simulation results of other typical attacks, e.g., JPEG compression and blurring, as an approximation.
The experimental results on the dataset YouTube-VOS demonstrate that our watermarking scheme simultaneously achieves satisfactory imperceptibility and robustness, and the accuracy of tampering localization is much higher compared to existing passive forensics methods.

Our contributions are mainly as follows: 1) We use deep networks for video watermarking against tampering; 2) We propose a tailored attacking layer for enhanced robustness against typical video post-processing attacks; 3) Our method can achieve higher accuracy in locating tampered regions and is robust against multiple kinds of post-processing attacks.

\section{Method}
\subsection{Approach Overview}
\begin{figure} [t!]
	\centering
	\includegraphics[width=1.0\linewidth]{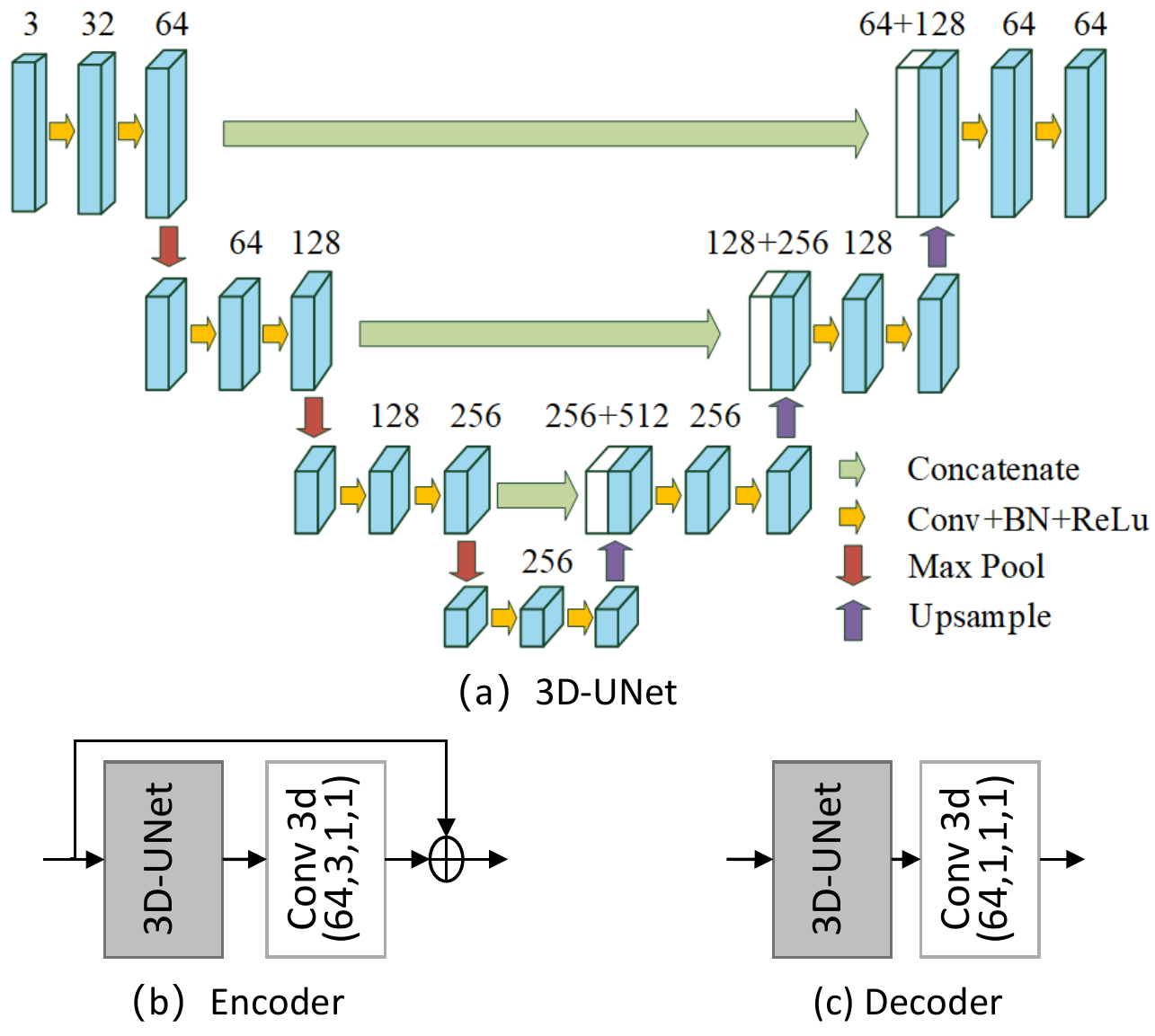}
	\caption{\textbf{Architectures of the encoder and decoder,} which are based on the 3D-Unet~\cite{cciccek20163d}.}
	\label{Fig2} 
\end{figure}

Fig.~\ref{Fig1} illustrates the network design of the proposed method.
Our watermarking scheme follows the traditional data hiding pipeline, which mainly contains three phases, namely, watermark embedding, attacking simulation and forgery detection. We use two independent three-dimensional U-shaped architecture~\cite{cciccek20163d} to hide auto-generated watermark into an original video, and localize the tampered areas on receiving the attacked version, respectively. 
In detail, given an original video $\mathbf{V}$, we transform the original video $\mathbf{V}$ into the watermarked video $\mathbf{V}_E$ using the encoder. The attacking layer performs both tampering and benign video post-processing attacks on $\mathbf{V}_E$ to generate $\mathbf{V}_{\emph{ATK}}$. In this stage, the hidden information might be globally or locally destroyed. On the recipient's side, the decoder produces the predicted tampering mask $\hat{\mathbf{M}}$ to see which parts of the video are tampered with. The architectures of the encoder and decoder are shown in Fig.~\ref{Fig2}.

The objective functions include the embedding loss $\mathcal{L}_{\emph{emb}}$ and the localization loss $\mathcal{L}_{\emph{loc}}$. We respectively employ the Mean Squared Error (MSE) loss and the Binary Cross-Entropy (BCE) loss as their implementation. 
\begin{equation}
\mathcal{L}_{\emph{emb} }=\|\mathbf{V}-\mathbf{V}_E\|_{2},
\end{equation}
\begin{equation}
\mathcal{L}_{loc}=-(\mathbf{M} \log \hat{\mathbf{M}}+(1-\mathbf{M}) \log (1-\hat{\mathbf{M}})),
\end{equation}
where $\mathbf{M}$ represents the groud-truth tampering mask.
The total loss is listed in Eq.~(\ref{eqn_loss_sum}), where $\alpha$ is a hyper-parameter.
\begin{equation}
\mathcal{L}=\mathcal{L}_{\emph{emb}}+\alpha\cdot \mathcal{L}_{\emph{loc}}.
\label{eqn_loss_sum}
\end{equation}

\subsection{Attack Simulation}
Attack simulation plays a critical role in robustness training. 
To simulate the video redistribution stage, we first perform tampering on $\mathbf{V}_E$, and afterwards, common video post-processing attacks are performed to generate the attacked video $\mathbf{V}_{\emph{ATK}}$.
To begin with, we select some critical areas within the original video $\mathbf{X}$ to form $\mathbf{M}$. 
In our scheme, we use binarized segmentation masks as $\mathbf{M}$. 
We then tamper the watermarked video $\mathbf{V}_E$ as $\mathbf{V}_{\emph{tmp}}$ by replacing contents within $\mathbf{M}$ with that within $\mathbf{R}$, a randomly-selected video frame from an irrelevant video. 
Next, we implement typical video post-processing attacks using $\emph{IP}(\cdot)$ to simulate that $\mathbf{V}_{\emph{tmp}}$ must be lossily processed during transmission and storage, where the attacker wants to conceal the tampering behavior. In sum, the attacked video $\mathbf{V}_{\emph{ATK}}$ is generated according to Eq.~(\ref{eqn_tampering}).
\begin{equation}
\mathbf{V}_\emph{ATK}=\emph{IP}(\mathbf{X} \cdot(1-\mathbf{M})+\mathbf{R} \cdot \mathbf{M}). 
\label{eqn_tampering}
\end{equation}

The video post-processing attacks implemented in our scheme, i.e., $\emph{IP}(\cdot)$, include the following attacks. (1) \textbf{Median filtering}, where each output pixel is computed as the median value of the input pixels under a $5\times~5$ window. (2) \textbf{Gaussian blurring}, which convolves the image with a Gaussian $5\times~5$-sized kernel. (3) \textbf{Rescaling}, which randomly scales the video frames up or down and back to the original size. (4) \textbf{Image lossy compression}, which lossily compresses the video frame by frame. (5) \textbf{Video compression and decompression (codec)}, which lossily compresses the video using both spatial and temporal characteristics of the video.
These attacks might introduce missing or distorted higher-frequent details in the ultimate videos compared to the original ones. For example, JPEG compression is a widely used compression that includes DCT transformation, quantization on the coefficients and lossless encoding, in which the quantization process discards many details for file size shrinkage, and can cause chess-board artifacts. Video codec attack is similar yet more complicated than the other attacks. The decompressed video might have lower quality than the original video because there is insufficient information to accurately reconstruct the original video. 
Typical video codec standards are MPEG, H.264, HEVC (H.265), etc.

\begin{figure*}[!t]
\centering
\includegraphics[width=0.9\linewidth]{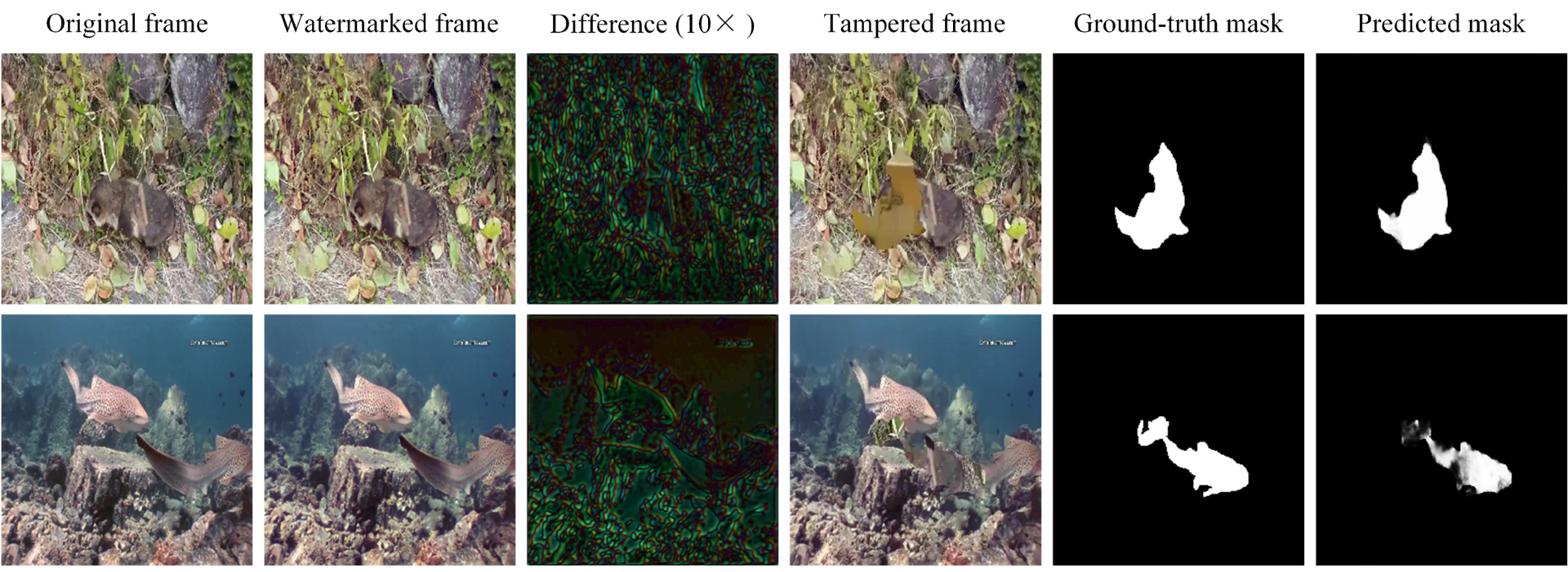}
\caption{\textbf{Performance against tampering with FFMPEG compression ($\rm{CRF} = 17$). The average PSNR and SSIM on YouTube-VOS dataset are 37.78 dB and 0.987, respectively.} The difference between the original video and the watermarked video cannot be perceived by the human eye, and the tampered area can still be accurately detected in the compressed video.}
\label{Fig3}
\end{figure*}

The issue mainly lies in how to effectively simulate video codec attack, since the leading three kinds of attacks can be easily implemented by simple differentiable methods~\cite{zhu2018hidden}. As for image compression, many effective JPEG simulators have been proposed in the past literature, such as Diff-JPEG~\cite{shin2017jpeg}, MBRS~\cite{jia2021mbrs} and HiDDeN~\cite{zhu2018hidden}.  
In contrast, there is no off-the-shelf differentiable video codec simulator so far. The reason is that many internal steps, such as motion estimation and compensation, are hard to be differentiated. 
However, we find that JPEG and H.264 compression both share the process of DCT-based coefficient quantization, suggesting that losses introduced by video codec might share some common characteristics with those made by the rest of the attacks.
Therefore, we are motivated to design a video codec simulator by ensembling the attacked videos generated by blurring, scaling, JPEG compression, etc. 
First, we use a real video codec to compress $\mathbf{V}_\emph{ATK}$ according to different coding standards and Constant Rate Factors (CRFs), i.e., $\mathbf{V}_\emph{ATK}^{codec}=\emph{Real\_Codec}(\mathbf{V}_\emph{tmp},\emph{CRF})$, where $\emph{Real\_Codec}$ is the H.264 codec, and $\emph{CRF}=\{17,23,29\}$. Note that CRF is a tunable content-specific offset to the frame’s quantization parameter, with lower values indicating less compression and higher quality. We select the H.264 codec for its overwhelming popularity.
The attacked video by the video codec attack simulation is generated according to Eq.~(\ref{eqn_codec_simulation}).
\begin{equation}
 \begin{split}
\hat{\mathbf{V}}_\emph{ATK}^{\emph{codec}}=&
\alpha_{0}\cdot\emph{Resize}(\mathbf{V}_\emph{tmp})
+\alpha_{1}\cdot\emph{Med}(\mathbf{V}_\emph{tmp})\\
&+\alpha_{2}\cdot\emph{Gauss}(\mathbf{V}_\emph{tmp})
+\alpha_{3}\cdot\emph{JPEG}(\mathbf{V}_\emph{tmp},\emph{QF}_W)\\
&+\alpha_{4}\cdot\emph{JPEG}(\mathbf{V}_\emph{tmp},\emph{QF}_S),
 \end{split}
\label{eqn_codec_simulation}
\end{equation}
where $\emph{QF}_S\in\{40,50,60\}$ and $\emph{QF}_W\in\{70,80,90\}$, respectively represent strong and weak JPEG compression attack. $\alpha=\{\alpha_{0},...,\alpha_{4}\}$ are learnable parameters that weight the generated results of the five attacks to let $\mathbf{V}_\emph{ATK}$ more close to $\mathbf{V}_\emph{GT}$.
For $\mathbf{V}_\emph{ATK}^{codec}$ with totally three different combinations of codec and CRF, we employ three different sets of parameters $\alpha$.
On training the simulator, given a $\mathbf{V}_\emph{tmp}$, we randomly sample a codec and CRF and generate $\mathbf{V}_\emph{ATK}^{codec}$, we let the simulator update the corresponding $\alpha$ to make closer $\hat{\mathbf{V}}_\emph{ATK}^{codec}$ and $\mathbf{V}_\emph{ATK}^{codec}$. We use the MSE loss $\mathcal{L}_{\emph{simul}}$ as the supervision on the simulator.
\begin{equation}
\mathcal{L}_{\emph{simul} }=\|\hat{\mathbf{V}}_\emph{ATK}^{codec}-\mathbf{V}_\emph{ATK}^{codec}\|_{2}.
\end{equation}

Then, in the training phase, the benign attack $\emph{IP}(\cdot)$ is evenly and iteratively switched within the range of the above attacks. 
Empirically, we find that robustness against video codec simulation is much more important than robustness against the rest of the attacks. Therefore, we further propose two strategies in our adversarial training mechanism. First, we observe that the blurring results produced by median filtering and Gaussian blurring are close. We again use the ensembling strategy to linearly combine $\emph{Gauss}(\mathbf{V}_\emph{tmp})$ with $\emph{Med}(\mathbf{V}_\emph{tmp})$ as the mixed filtering operation. The benefit is that we moderately lower the importance of robustness against blurring, and further introduce randomness within the model.
Second, in some cases, we directly let $\hat{\mathbf{V}}_\emph{ATK}^{codec}=\mathbf{V}_\emph{ATK}^{codec}$, and address the non-differentiable problem by using the noise-addition strategy proposed by Zhang et al.~\cite{zhang2021towards}. That is, the residual $e=\mathbf{V}_\emph{ATK}^{codec}-\mathbf{V}_\emph{tmp}$ is detached and directly added onto $\mathbf{V}_\emph{tmp}$. Therefore, $\mathbf{V}_\emph{ATK}$ in different training iteration $\emph{iter}$ is as follows.
\begin{equation}
\mathbf{V}_\emph{ATK}=
\begin{cases}
\emph{Resize}(\mathbf{V}_\emph{tmp}), & \emph{iter}\%4=0 \\
\beta\cdot\emph{Med}(\mathbf{V}_\emph{tmp})+\gamma\cdot\emph{Gauss}(\mathbf{V}_\emph{tmp}), & \emph{iter}\%4=1 \\
\hat{\mathbf{V}}_\emph{ATK}^{\emph{codec}}, & \emph{iter}\%4=2 \\
\emph{stop\_grad}(e)+\mathbf{V}_\emph{tmp}, & \emph{iter}\%4=3 \\
\end{cases},
\end{equation}
where $\beta,\gamma\in[0,1],\beta+\gamma=1$. The reason for using hybrid real-world and simulated video codec is to reduce temporal complexity, where real video codecs are much slower than the proposed video simulator.

\begin{center}
\begin{table} [!t]
\centering
  \caption{\noindent\textbf{Results of the learned parameters of the codec simulator under different attacks.} Compression-based attacks are more preferred than the rest kinds of attacks. The highest and lowest value are respectively marked red and blue.}
  \label{simulator}
  \renewcommand{\arraystretch}{1.2}{
\begin{tabular}{ccccccc}
\toprule
&  \rm{CRF}  & Resize & \thead{Strong\\ JPEG} & \thead{Weak\\JPEG}  & \thead{Meadian\\Blurring}& \thead{Gaussian\\Blurring}\\
\midrule
 &17 & -0.7575&1.0723&\redmarker{1.4917}&\bluemarker{-1.3091}&-0.8802\\
 &23 &-0.7568&1.0685&\redmarker{1.3155}&-0.8185&\bluemarker{-1.1666}\\
 &29 & -0.1529&\redmarker{1.1229}&0.5848&-0.2182&\bluemarker{-0.7920}\\
    \bottomrule
  \end{tabular}}
\end{table}
\end{center}
\begin{figure*}[!t]
\centering
\includegraphics[width=0.9\linewidth]{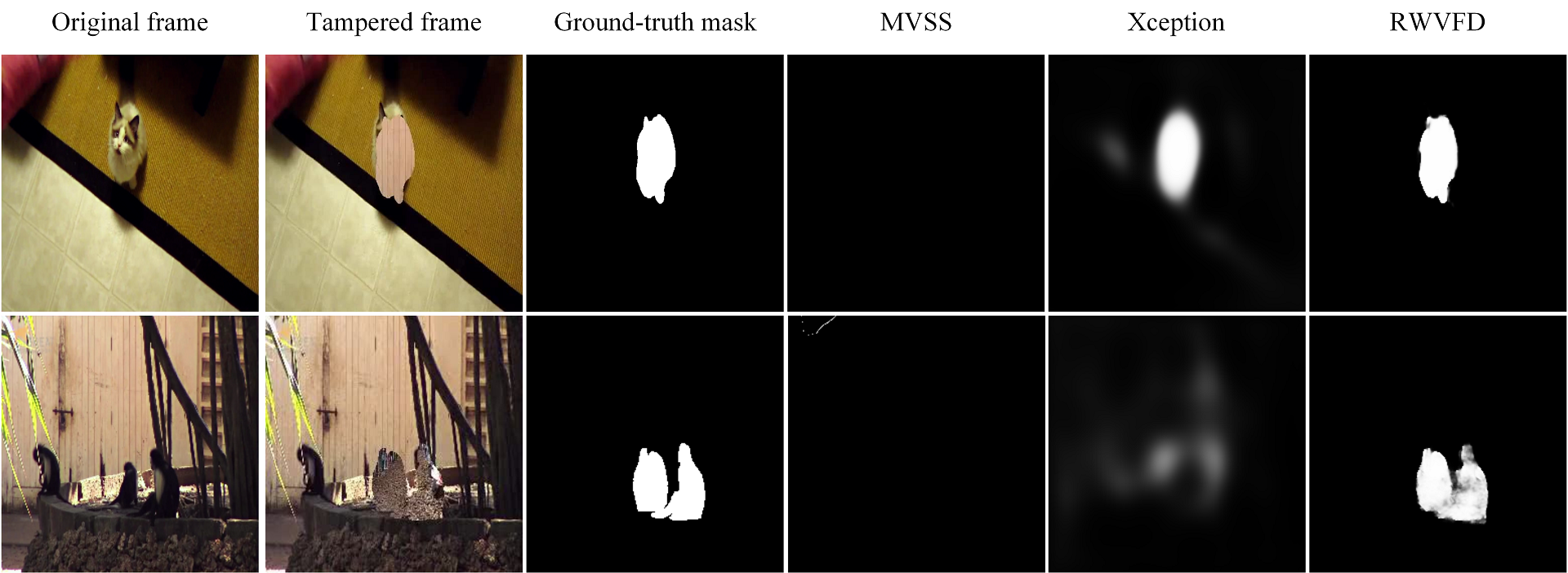}
\caption{\textbf{Comparison of tamper localization under FFMPEG compression ($\rm{CRF} = 17$) among RWVFD and several state-of-the-art passive schemes.} RWVFD can accurately localize the tampered areas even with the presence of post-processing attack. In contrast, many passive schemes have low detection accuracy and are reported to not have robustness.}
\label{Fig4}
\end{figure*}
\begin{table*} [!t]
  \caption{\noindent\textbf{Performance comparison  for tamper detection among our scheme and the state-of-the-art passive methods.} NA: No-attack, GB: Gaussian Blur, MB: Median Blur, VC: Video Compression, FD: Frame Dropping, FRC: Frame Rate conversion, CRF: Constant Rate Factor, DN: Dropping Number, FR: Frame Rate.}
  \label{comparison}
  \setlength{\tabcolsep}{3mm}
  \renewcommand{\arraystretch}{1.2}{
\begin{tabular}{cccccccccccc}
\toprule
\multicolumn{2}{c}{Attack} & \multicolumn{3}{c}{MVSS~\cite{dong2021mvss}} & \multicolumn{3}{c}{Xception~\cite{rossler2019faceforensics++}} & \multicolumn{3}{c}{RWVMD}\\

 \cline{3-11}&& Precision  & Recall & F1-score & Precision  & Recall & F1-score& Precision  & Recall & F1-score\\
\midrule
 \multicolumn{2}{c}{NA}  & 0.05& 0.00& 0.01& 0.64& 0.59&0.57&1.00& 0.99&0.99\\
 \multicolumn{2}{c}{GB}  &0.06& 0.00& 0.01& 0.55& 0.21&0.25& 0.99& 0.95&0.96\\
 \multicolumn{2}{c}{MB}  &0.05& 0.00& 0.01& 0.64& 0.60& 0.58& 0.99& 0.99&0.99\\
\cline{1-11}
\multirow{3}{*}{VC}  
&$\rm{CRF}$ = 17 &0.05& 0.00& 0.01& 0.66& 0.46& 0.49& 0.95&0.83&0.87\\
&$\rm{CRF}$ = 23 & 0.05& 0.00& 0.01& 0.67& 0.45& 0.49& 0.95&0.80&0.84\\
&$\rm{CRF}$ = 29 & 0.05& 0.00& 0.01& 0.68& 0.42& 0.46&  0.90& 0.70&  0.75\\
\cline{1-11}
\multirow{2}{*}{FD}  
& $\rm{DN}=1$ &0.05& 0.00& 0.01& 0.64& 0.59& 0.57& 0.93& 0.98&  0.95\\
& $\rm{DN}=2$ &0.05& 0.00& 0.01& 0.64& 0.59& 0.57& 0.97& 0.99&  0.97\\
\cline{1-11}
\multirow{3}{*}{FRC} 
&$\rm{FR} = 20$ & 0.05& 0.00& 0.01& 0.67 &0.46&0.49& 0.96& 0.92&  0.93\\
&$\rm{FR} = 30$ & 0.05& 0.00& 0.01& 0.66& 0.55& 0.55& 0.87& 0.87&  0.84\\
&$\rm{FR} = 35$ &0.05& 0.00& 0.01& 0.66& 0.55& 0.55& 0.88& 0.86&  0.84\\

    \bottomrule
  \end{tabular}}
\end{table*}

\section{Experiments}
\subsection{Experimental Setup}
We empirically set the hyper-parameter as $\alpha=0.9$. The batch size is set as four. We use Adam optimize~\cite{kingma2014adam} with the default parameters. The learning rate is $1 \times 10^{-4}$ with manual decay. The default frame rate of the video is set to 25 frames per second. We binarize the prediction mask by setting the threshold $\emph{Th}$ as $0.5$.

We use two popular object segmentation datasets, namely, Davis~\cite{perazzi2016benchmark} and YouTube-VOS~\cite{xu2018youtube} in the experiment. During training, the original videos $\mathbf{V}$ are prepared by selecting the data from the whole Davis dataset and YouTube-VOS train set. We use the YouTube-VOS test set to test our model. The tampering masks are the annotation images corresponding to the video frames in the object segmentation datasets.

We compare RWVFD with two passive methods for tamper detection, which detect universal manipulations or deepfake, namely, MVSS-Net~\cite{dong2021mvss} and Xception~\cite{rossler2019faceforensics++}.
We employ the peak signal-to-noise ratio (PSNR) and the Structural Similarity~\cite{wang2004image} to evaluate the image quality. The value of SSIM ranges from zero to one. A higher structure similarity is indicated by a high SSIM closer to one. 
We employ the Precision, Recall, and F1 score to measure the accuracy of tamper localization.
Higher F1 value indicates more accurate result.

\subsection{Imperceptibility of watermark embedding}
In Fig.~\ref{Fig3}, we showcase the first frames from three randomly selected test videos from YouTube-VOS dataset. We can observe that the differences before and after watermark embedding are almost imperceptible, and the overall quality of the watermarked frames is satisfactory. The watermark information is distributed in the whole set of video frames, mainly hidden in the higher frequencies. The embedded watermark is robust to video processing attacks and can be used for tampering localization. Instead of finding a ubiquitously existing trace to unveil video modification behavior, in our scheme, tampering will result in local pattern inconsistencies, allowing the network to efficiently detect and locate the tampered regions. We have conducted the embedding experiments on the entire test dataset of YouTube-VOS and the average PSNR and SSIM are 37.78 dB and 0.987, respectively.

\subsection{Accuracy and robustness of tampering localization}
In Table~\ref{comparison}, we clarify the robustness of RWVFD with the presence of different video processing attacks.  It can be seen from the results that RWVFD has strong robustness to common video processing behaviors. Even if there is a high-intensity H.264 video compression attack, the performance will not be significantly degraded. It proves the effectiveness of our codec simulator for improving watermarking robustness.  From Fig.~\ref{Fig3}, RWVFD can accurately detect forged regions even under compression attacks. In addition, RWVFD is also robust to typical temporal attacks such as frame deletion and frame rate transformation. 

Table~\ref{simulator} shows the learned parameters of the video codec simulator. On simulating a video codec, the simulator prefers the simulated images of JPEG compression, and those of blurring attacks will be suppressed, indicating that the video coding distortion shares more characteristics with two-dimensional JPEG compression, such as chess-board artifact.

In comparison, passive forensics methods, e.g., MVSS and Xception, do not perform well on the test sets. We use the pre-trained models provided by the authors of MVSS, and find that the model fails to detect forged regions when there exist video codec attacks. The reason is most likely that video frames after codec attack are out of distribution of natural images that MVSS detects. We train Xception using the same training dataset as RWVFD. We also equip the scheme with the attacking layer proposed by us for adversarial training. However, the accuracy is still not high. Fig.~\ref{Fig4} showcases the experimental comparison of tamper localization among RWVFD, MVSS, and Xception.

\begin{table*} [!t]
  \caption{\noindent\textbf{Ablation study of RWVFD using varied partial settings.}}
  \label{Ablation}
  \setlength{\tabcolsep}{3mm}
  \renewcommand{\arraystretch}{1.2}{
\begin{tabular}{cccccccccccc}
\toprule
 ~&~& \multicolumn{3}{c}{RWVMD w/o 3D-Unet } & \multicolumn{3}{c}{RWVMD w/o codec simulator} & \multicolumn{3}{c}{RWVMD} \\
\multicolumn{2}{c}{Attack}&\multicolumn{3}{c}{$(\rm{PSNR}=34.95 dB, SSIM=0.958)$}&\multicolumn{3}{c}{$(\rm{PSNR}=35.82 dB, SSIM=0.987)$}&\multicolumn{3}{c}{$(\rm{PSNR}=37.78 dB, SSIM=0.987)$}\\

 \cline{3-11}&~& Precision  & Recall & F1-score & Precision  & Recall & F1-score& Precision  & Recall & F1-score\\
\midrule

\multirow{3}{*}{VC}  
&$\rm{CRF}$ = 17 &0.72& 0.94& 0.80& 0.76& 0.86& 0.80& 0.95&0.83&0.87\\
&$\rm{CRF}$ = 23 & 0.66& 0.94& 0.75& 0.71& 0.87& 0.77& 0.95&0.80&0.84\\
&$\rm{CRF}$ = 29 & 0.49& 0.95& 0.61& 0.44& 0.89& 0.56&  0.90& 0.70&  0.75\\
\cline{1-11}
\multirow{2}{*}{FD}  
& $\rm{DN}=1$ &0.98& 0.98& 0.97& 0.44& 0.99& 0.56& 0.93& 0.98&  0.95\\
& $\rm{DN}=2$ &0.98& 0.98& 0.97&0.49& 0.99&0.61& 0.97& 0.99&  0.97\\
\cline{1-11}
\multirow{3}{*}{FRC} 
&$\rm{FR} = 20$ &0.90& 0.96& 0.92& 0.31& 0.97&0.42& 0.96& 0.92&  0.93\\
&$\rm{FR} = 30$ &0.90& 0.96& 0.93& 0.21& 0.99& 0.31& 0.87& 0.87&  0.84\\
&$\rm{FR} = 35$ &0.91& 0.96& 0.92& 0.21& 0.99& 0.31& 0.88& 0.86&  0.84\\

    \bottomrule
  \end{tabular}}
\end{table*}

\subsection{Ablation Study}
We explore the influence of 3D-Unet and codec simulator in our scheme. In the first experiment, we changed the 3D-Unet architecture into 2D-Unet as RWVMD without 3D-Unet. In the second experiment, we train the network without using the codec simulator. For fair comparisons, we separately train the model from scratch until it converges, perform the same video post-processing attacks and the same tampering attack in each test. 

We summarize the average results on the test dataset in Table~\ref{Ablation}. The results show that the complete implementation of RWVFD has higher PSNR and F1 scores than the ablated versions.
In comparison, RWVMD without 3D-UNET cannot perform forgery detection when we apply video codec attacks. It suggests that considering video frames as independent images and embedding temporarily-inconsistent watermarks is less effective in countering typical video attacks. 
RWVMD without the proposed codec simulator also performs worse in the overall accuracy under video compression. This shows that applying Zhang et al.~\cite{zhang2021towards} alone is not enough, and proves the necessity of applying the codec simulator. 

\section{Conclusion}
In this paper, we propose a deep learning-based video watermarking method RWVFD for video forgery detection. We encode the original video into a watermarked video, in which the tampering attack area can be accurately located. To improve performance, we propose a  video codec simulator along with simulation of other typical attacks to enhance the robustness against common video post-processing attacks. We conduct experiments on a popular video dataset and the results demonstrate the effectiveness of RWVFD in tampering localization.


\bibliographystyle{IEEEbib}
\bibliography{reference}

\end{document}